\documentclass[letterpaper]{article} 
\usepackage{aaai2026}  
\usepackage{times}  
\usepackage{helvet}  
\usepackage{courier}  
\usepackage[hyphens]{url}  
\usepackage{graphicx} 
\usepackage{booktabs} 
\usepackage{siunitx} 
\usepackage{multirow} 
\usepackage{amsmath}
\usepackage{amsfonts}
\usepackage[sort]{natbib}
\usepackage{natbib}  
\usepackage{caption} 
\usepackage{algorithm}
\usepackage{algorithmic}

\usepackage{newfloat}
\usepackage{listings}
\DeclareCaptionStyle{ruled}{labelfont=normalfont,labelsep=colon,strut=off} 
\floatstyle{ruled}
\newfloat{listing}{tb}{lst}{}
\floatname{listing}{Listing}
\nocopyright

\title{Contrastive Cross-Bag Augmentation for Multiple Instance Learning-based Whole Slide Image Classification}
\author{
    Bo Zhang\textsuperscript{\rm 1}, 
    Xu Xinan\textsuperscript{\rm 1}, 
    Shuo Yan\textsuperscript{\rm 1}, 
    Yu Bai\textsuperscript{\rm 1}, 
    Zheng Zhang\textsuperscript{\rm 2}, 
    Wufan Wang\textsuperscript{\rm 1}, 
    Wendong Wang\textsuperscript{\rm 1}
}
\affiliations{

    \textsuperscript{\rm 1} State Key Laboratory of Networking and Switching Technology, School of Computer Science (National Pilot Software Engineering School), Beijing University of Posts and Telecommunications.\\
    \textsuperscript{\rm 2} School of Intelligent Engineering and Automation, Beijing University of Posts and Telecommunications.\\
    zbo@bupt.edu.cn, xuxinan@bupt.edu.cn, shuoyan@bupt.edu.cn, by@bupt.edu.cn, zhangzheng@bupt.edu.cn, wufanwang@bupt.edu.cn, wdwang@bupt.edu.cn
}

\usepackage{bibentry}

\begin{document}

\maketitle

\begin{abstract}
Recent pseudo-bag augmentation methods for Multiple Instance Learning (MIL)-based Whole Slide Image (WSI) classification sample instances from a limited number of bags, resulting in constrained diversity.
To address this issue, we propose Contrastive Cross-Bag Augmentation ($\text{C}^{2}$Aug) to sample instances from all bags with the same class to increase the diversity of pseudo-bags.
However, introducing new instances into the pseudo-bag increases the number of critical instances (e.g., tumor instances).  
This increase results in a reduced occurrence of pseudo-bags containing few critical instances, thereby limiting model performance, particularly on test slides with small tumor areas.
To address this, we introduce a bag-level and group-level contrastive learning framework to enhance the discrimination of features with distinct semantic meanings, thereby improving model performance.
Experimental results demonstrate that $\text{C}^{2}$Aug consistently outperforms state-of-the-art approaches across multiple evaluation metrics.
\end{abstract}


\section{Introduction}


Whole Slide Image (WSI) classification aims to assign a class label, such as cancer type, to each slide.  
Each slide has a high resolution, typically on the order of $100000 \times 100000$~\cite{transmil}.  
Due to this high resolution, manually annotating tumor regions within each slide is a time-consuming task~\cite{ZHANG2025103027_wsi_survey}.
%
%
Therefore, numerous studies have adopted Multiple Instance Learning (MIL) for WSI classification~\cite{transmil, dtfd, dsmil}.
These methods first extract patches from the WSI and then encode them into patch-level features, which are treated as instances.
All instances derived from the same slide are considered as a bag, and only the bag-level label is required during training.
MIL-based approaches enable WSI classification without the need for tumor annotation.
However, WSIs are difficult to access publicly as they are typically stored in the data centers of individual hospitals.
The limited size of datasets for WSI classification constrains the performance of MIL-based methods.

\begin{figure}[!t]
  \centering
  \includegraphics[width=1\linewidth]{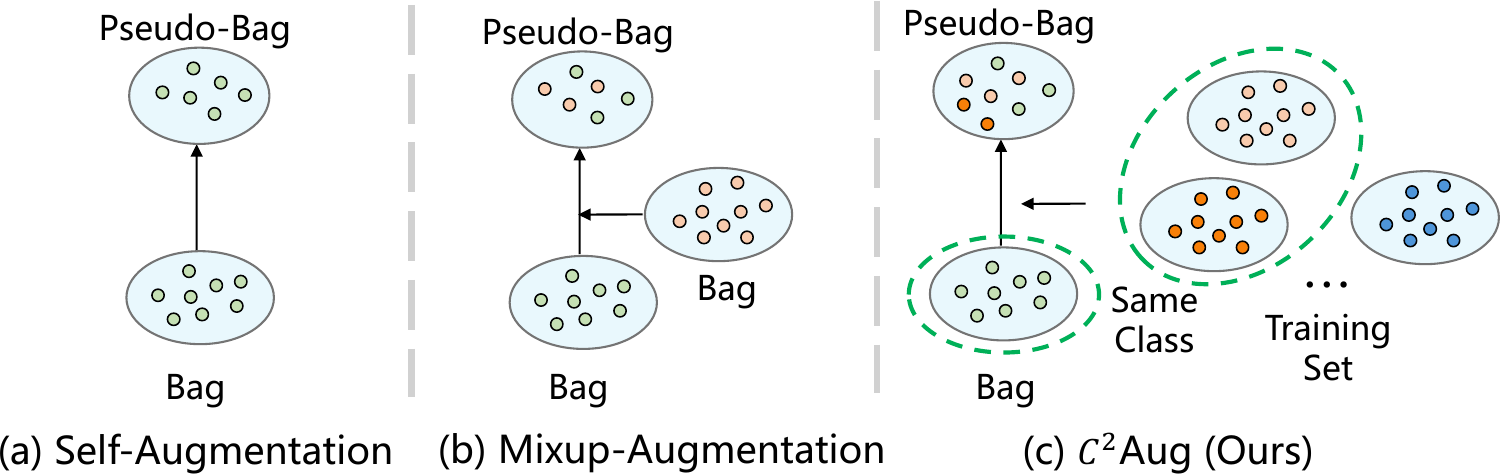}
  \vspace{-1.4em}
  \caption{
    Motivation of $\text{C}^{2}$Aug.
    (a) Self-augmentation generates a pseudo-bag by sampling instances within the input bag.
    (b) Mixup-style augmentation creates a pseudo-bag by merging two distinct bags.
    (c) $\text{C}^{2}$Aug generates a pseudo-bag by sampling instances from all bags sharing the same bag-level label.
  }
  \label{fig:fig1}
  \vspace{-1.6em}
\end{figure}

Recent studies~\cite{psemix, rankmix, mixupmil, remix} have proposed various MIL-based data augmentation techniques to increase the number of training samples.
These methods can be roughly classified into two categories: self-augmentation and Mixup-style augmentation methods.
Self-augmentation methods~\cite{augdiff, zaffar2023embedding_DA-GAN, tang2024self_ssrdl} generate pseudo-bag by augment instances within the input bag.
Mixup-style augmentation methods merge two bags into a single pseudo-bag by sampling instances from these two bags, inspired by the Mixup approach~\cite{zhang2017mixup_mixup}.
These two categories of methods generate pseudo-bags based on a limited number of bags.  
However, this constraint limits the diversity of the resulting pseudo-bags, thereby hindering the performance of MIL models.

To alleviate this limitation, we propose $\text{C}^{2}$Aug, which samples instances from all bags sharing the same bag-level label and combines them with the input bag to generate pseudo-bags, as illustrated in Fig.~\ref{fig:fig1}.
$\text{C}^{2}$Aug consists of three augmentation strategies: Multi-View Fusion, which performs instance-level augmentation; and Instance Compression and Expansion, which aim to augment the bag size by altering the number of instances.
Multi-View Fusion samples instances from all bags 
sharing the same class-level label 
and fuses them with the input bag to increase the diversity of each instance.
Instance Expansion performs dataset-level sampling and concatenates the sampled instances with the input bag to increase the bag size.
Instance Compression employs Cross Attention with a randomly sampled compression ratio to reduce the size of the input bag.
Unlike recent augmentation methods that discard instances from the input bag~\cite{psemix, rankmix},  
$\text{C}^{2}$Aug generates pseudo-bags by incorporating new instances into the original input bag,  
which helps reduce label noise~\cite{dong2025disentangled_DPBAug}.
However, this increase in tumor instances within pseudo-bags leads to a reduced occurrence of pseudo-bags containing only a small number of tumor instances during training.
This is particularly problematic for datasets containing a small tumor area~\cite{cam16}.
Therefore, we further propose Bag-level and Group-level Contrastive Learning to bring semantically similar features closer in the embedding space, thereby facilitating more accurate classification by the MIL model.
Bag-level contrastive learning employs a two-branch (student and teacher) framework to perform contrastive learning.
Since instances with similar features often appear in different bags~\cite{wang2023retccl_cross_level_group}, directly applying instance-level contrastive learning may introduce noise.
To address this, we cluster instances with similar features into the same groups.  
We then perform size alignment by compressing the instances within each group and subsequently apply Group-level Contrastive Learning.

Our main contributions are as follows:  
(1) We propose a Cross-Bag Augmentation module comprising three augmentation methods for compression-based instance and bag-level augmentation.  
(2) We design a Bag-level Contrastive Learning method to bring bag-level representations with similar semantics closer in the embedding space.  
(3) We introduce a Group-level Contrastive Learning method to encourage instances with similar features to cluster together.  


\section{Related Work}
\noindent\textbf{Pseudo-bag Augmentation for MIL based WSI classification.} 
MIL-based WSI classification methods~\cite{abmil, dtfd, dsmil, transmil} are widely adopted in this field.
However, due to the limited number of training samples, the performance of MIL models is constrained.
Therefore, some methods~\cite{dong2025disentangled_DPBAug, augdiff, psemix, rankmix} propose to generate pseudo-bags to increase the training samples.
Recent augmentation methods can be categorized into two classes: Self-Augmented and Mixup-Augmented methods.
Self-Augmented methods generate pseudo-bags from a single bag.
Augdiff~\cite{augdiff} and DA-GAN~\cite{zaffar2023embedding_DA-GAN} adopts generative models to perform instance-level augmention.
PRDL~\cite{25aaai} leveraging prompts to guide data augmentation in feature space.
Mixup-Augmention~\cite{rankmix, psemix, remix, mixupmil} methods generate pseudo-bags by mixing two bags.
RankMix~\cite{rankmix} samples instances from two bags based on the attention score.
DBPAug~\cite{dong2025disentangled_DPBAug} extends instance sampling to four bags, drawing an equal number of instances from each bag.
However, DBPAug cannot flexibly adjust the number of instances per bag and may discard critical instances during augmentation, as it samples instances based on instance-level prediction scores.


\noindent \textbf{WSI-based Contrastive Learning.}
Recent WSI classification methods~\cite{wang2023retccl_cross_level_group, dsmil, 25aaai, wang2022scl_scl_wc, zhu2022murcl_murcl} adopt contrastive learning such as MoCo~\cite{moco}, SimCLR~\cite{chen2020simple_simclr} to improve the model performance.
Some methods focus on improve the patch-level representation~\cite{dsmil, 25aaai}.
MuRCL~\cite{zhu2022murcl_murcl} leverages reinforcement learning to dynamically select discriminative feature sets for contrastive learning.
SCL-WC~\cite{wang2022scl_scl_wc} performs bag-level contrastive learning to enhance the discriminability of bags from different classes.
RetCCL~\cite{wang2023retccl_cross_level_group} adopts both bag-level and group-level contrastive learning to improve the model performance.
However, RetCCL employs K-means to compute cluster centers rather than using learnable prototypes; consequently, the cluster centers can only naively aggregate instances within each group in an untrainable manner, limiting their representational capacity.

\begin{figure*}[!t]
  \centering
  \includegraphics[width=0.95\textwidth]{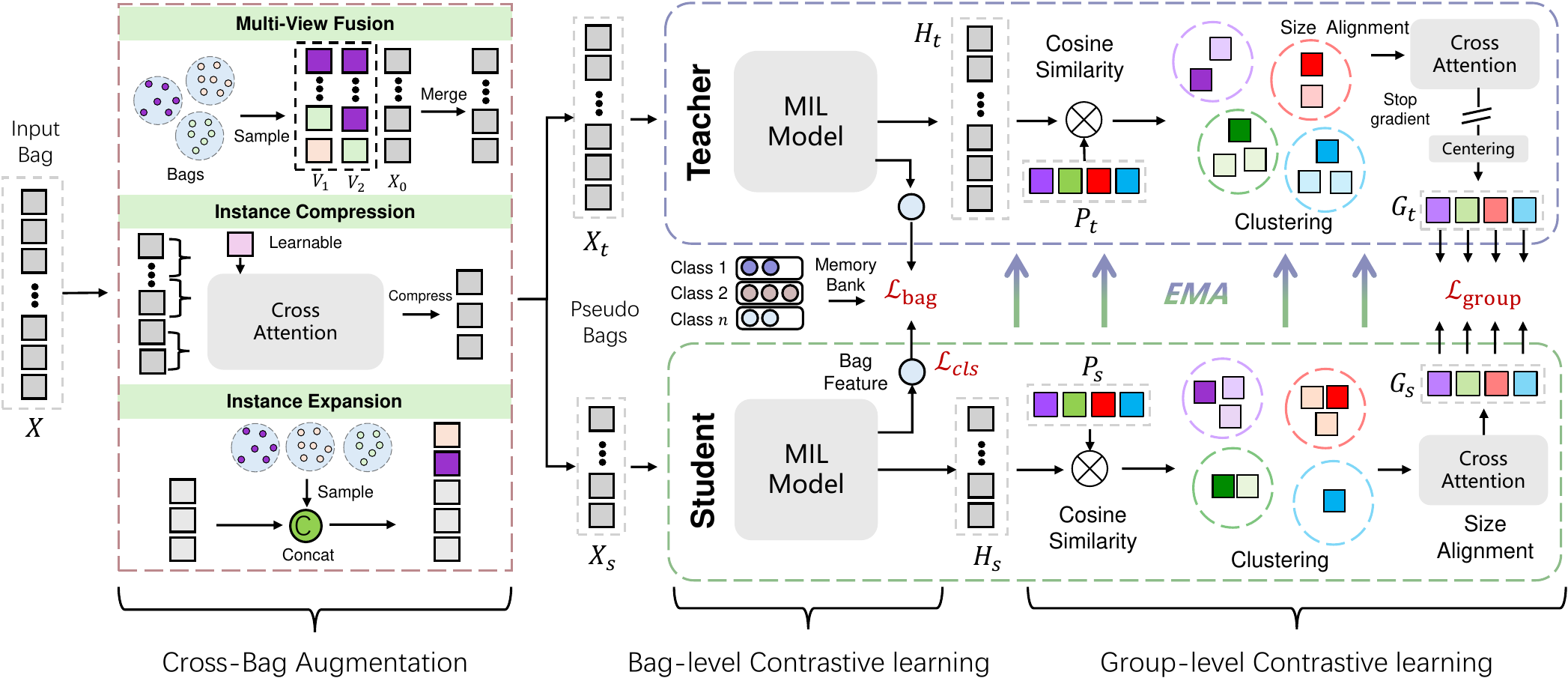}
  \vspace{-1.0em}
  \caption{
    $\text{C}^{2}$Aug consists of three steps.
    (1) \textbf{Cross-Bag Augmentation}
    adopts three augmentations to generate two different pseudo-bags from the same input bag, which are then fed into the teacher and student MIL models, respectively.
    (2) \textbf{Bag-Level Contrastive Learning}
   computes the bag-level contrastive loss $L_{bag}$ to bring bag representations with similar features closer in the embedding space.
    The parameters of the teacher model are updated using exponential moving average (EMA).
    (3) \textbf{Group-level Contrastive Learning}
    assigns each instance to a learnable prototype, performs size alignment by compressing the instances toward their assigned prototypes, and finally computes $L_{group}$.
  }
  \vspace{-1.3em}
  \label{fig:fig2}
\end{figure*}

\section{Method}
Given a bag $X_i$, $\text{C}^{2}$Aug generates a pseudo-bag by combining $X_i$ with instances sampled from other bags that share the same bag-level label, as illustrated in Fig~\ref{fig:fig2}.
$\text{C}^{2}$Aug first employs Cross-Bag Augmentation to generate two pseudo-bags (Section~\ref{sec:mult-bag}).
Then we feed these two pseudo-bags into the student and teacher MIL models to compute the WSI classification loss $L_{cls}$, along with the bag-level contrastive loss $L_{bag}$, which encourages semantically similar features to stay close to each other (Section~\ref{sec:bag-level}).
Finally, we perform group-level contrastive learning to further separate tumor and normal instances in the embedding spaces (Section~\ref{sec:bag-level}).
$\text{C}^{2}$Aug can generate more diversified pseudo-bags by sampling instances from all bags with the same bag-level label and push instances with different semantic features apart from each other in the embedding space.

\subsection{Background: MIL Formulation}
MIL considers each WSI as a bag $X_i$, with each bag containing multiple instances.
These instances are the patch features extracted from the WSI, denoted as $\{x_{i, 0}, x_{i, 1}, ..., x_{i, n}\}$, where $n$ is the number of instances in the bag and may vary across different bags.
A bag is labeled positive if at least one of its instances is positive, and negative only if all instances are negative, as: 
{\small
\begin{equation}
    Y_i =
    \begin{cases}
        1, & \text{if } \exists j \text{ s.t. } y_{i,j} = 1, \\
        0, & \text{otherwise}.
    \end{cases}
\end{equation}
}
where $\mathbf{x}_{i,j}$ is the $j$-th instance and $y_{i,j} \in \{0,1\}$ is its (unknown) label, the bag label $Y_i$ is defined as $Y_i = 1$ if $\exists j$ such that $y_{i,j} = 1$, otherwise $Y_i = 0$. 

\subsection{Cross-Bag Augmentation}
\label{sec:mult-bag}


Cross-bag Augmentation proposes three steps to increase the diversity of each pseudo-bag: Multi-View Fusion, Instance Compression, and Instance Expansion.
Multi-View Fusion aims to increase the diversity of each instance, while Instance Compression and Expansion are designed to increase the diversity of bag size.


\noindent\textbf{Multi-View Fusion.}
Multi-View Fusion samples instances from multiple bags and fuses them into the input bag instances to increase the instance diversity.
Given a bag $X_i$, Multi-View Fusion samples instances from all bags with the same bag-level label to form multiple views $V_1, V_2, ..., V_m$, each of which has the same number of instances as $X_i$.
Instances within one view do not appear in other views.
Then we fuse $m$ views into $X_i$, so that the size of $X_{i}$ does not change when generating pseudo-bag $X_{i}^{'}$.

Specifically, for the $j$-th instance $x_{i, j}$, we fuse all the $j$-th instances from $m$ views $v_{1, j}, v_{2, j}, ..., v_{m, j}$ through Cross-Attention~\cite{chen2021crossvit_crossattention, li2023blip_qformer}:
{\small
\begin{equation}
\text{CrossAttention}(\mathbf{Q}, \mathbf{K}, \mathbf{V}) = \text{Softmax}\left( \frac{\mathbf{Q} \mathbf{K}^T}{\sqrt{d}} \right) \mathbf{V},
\label{eq:cross_attn}
\end{equation}
}
where $\mathbf{Q}$ is obtained from $x_{i, j} \in \mathbb{R}^{d}$ via a linear transformation; $\mathbf{K}$ and $\mathbf{V}$ are obtained from $m$ instances across multiple views $\{v_{1, j}, v_{2, j}, ..., v_{m, j}\}$ with shape $\mathbb{R}^{m \times d}$; $d$ is the instance feature dimension.
Each instance $x_{i, j}$ is fused with $m$ instances from other bags sharing the same label.

To further enhance the diversity of instances, we set $m$ to different values with respect to different $j$, so that each instance in bag $X_i$ is fused with a varying number of instances from other bags.
We set the range of the new number of instances $m^{'}$ between $1$ and $m$.
To achieve this, we mask the attention matrix $\mathbf{Q}\mathbf{K}^{T}$ by setting masked entries to -9999, which results in zero attention weights after the Softmax operation.
As a result, the masked instances have no effect on the output, as shown in Equation~\ref{eq:cross_attn}.
For each instance $x_{i,j} \in \mathbb{R}^{d}$ used as $\mathbf{Q}$, and $m$ instances $\{v_{1, j}, v_{2, j}, ..., v_{m, j}\}$ used as $\mathbf{K}$, the $ \mathbf{Q}\mathbf{K}^{T} $ matrix has shape $ \mathbb{R}^{m } $, where $ 0 \leq j < n $.
We design three types of masking strategies as follows:
\begin{itemize}
    \item \textbf{Element-wise Random Masking:} 
    Each element in the $\mathbf{Q}\mathbf{K}^{T}$ matrix has the same probability of being masked, and the masking of each element is performed independently.
    The probability of the number of masked elements $m_{j}^{'}$ is: 
    {\small
    \begin{equation}
    P(X = m_{j}^{'}) = \binom{m}{m_{j}^{'}} p^{m_{j}^{'}} (1 - p)^{m- m_{j}^{'}},
    \label{label:element_wise_random_mask}
    \end{equation}
    }
    where $m_{j}^{'} = 0, 1, \ldots, m$, $p$ is the probability that masks each element.
    \item \textbf{Row-wise Random Masking:} 
    We randomly choose $m_{j}^{'}$, where $0 \leq m_{j}^{'} \leq m$, for each row in the $\mathbf{Q}\mathbf{K}^{T}$ matrix.
    Then, we randomly mask $m_{j}^{'}$ elements in each corresponding row of the $\mathbf{Q}\mathbf{K}^{T}$ matrix.
    The probability of the number of masked elements $m_{j}^{'}$ is:
    {\small
    \begin{equation}
        P(X = m_{j}^{'}) = \frac{1}{m}.
    \label{label:row_wise_random_mask}
    \end{equation}
    }
    \item \textbf{Top-$k$ Random Masking:}
    Similar to Row-wise Random Masking, we first randomly choose $m_{j}^{'}$ for each row in the $\mathbf{Q}\mathbf{K}^{T}$ matrix.
    Then, instead of randomly selecting elements, we select the top-$k$ elements in each row based on their attention scores, where $k$ equals $m_{j}^{'}$.        
    The probability of the number of masked elements is the same as Equation~\ref{label:row_wise_random_mask}.
\end{itemize}

Multi-View Fusion focuses on increasing the diversity of each instance by fusing instances from other bags.
By adopting this approach, it is able to increase the pseudo bag diversity while ensures that the pseudo-bag contains class-related features, whereas recent sampling-based pseudo-bag augmentation methods~\cite{dtfd, rankmix, psemix} cannot guarantee this.


\noindent \textbf{Instance Expansion.}
Instance Expansion samples instances from other bags to increase the bag size.
Specifically, we sample instances from all bags with the same bag-level label as $X_{i}$ to increase the bag size, and then concatenate them with $X_{i}$ to form a pseudo-bag. 
The number of sampled instances $n_{e} \in \mathbb{R}$ is drawn from a uniform distribution $[0, n_{max}-n]$, where $n$ is the size of bag $X_{i}$, and $n_{max}$ is the maximum bag size within the training set. 
After instance expansion, we obtain a pseudo-bag with bag size $n + n_{e}$. 
Instance Expansion samples instances from multiple bags to encode more patterns into $X_{i}$, while preserving the features relevant to the bag-level label.
\noindent\textbf{Instance Compression.}
We compress the instances within a bag to generate pseudo-bags with reduced bag size. 
Specifically, for an input bag $X_i \in \mathbb{R}^{n \times d}$, we first define a compression ratio $C_r$, and split all the instances into $C = \left\lceil\frac{n}{C_r}\right\rceil$ folds, each having $C_r$ instances. 
If $n$ cannot be divided by $C_r$, we pad $X_i$ with instances sampled from $X_i$. 
For the $c$-th fold with $C_r$ instances, we fuse the $C_r$ instances through a Cross-Attention module. 
We adopt a learnable weight as query $Q$ with shape $\mathbb{R}^{d}$ and the $C_r$ instances as input keys $K$ and values $V$ with shapes $\mathbb{R}^{C_r \times d}$ for Cross-Attention, resulting in a compressed instance $x_{i}^{c}$.
The learnable query is shared across all folds.
To increase diversity, $C_r$ is uniformly sampled from $[2, C_r]$ during each iteration.
Therefore, an input bag with $n$ instances will result in a pseudo-bag with $C$ instances. 
Compared to other pseudo-bag augmentation methods~\cite{psemix, rankmix} that use sampling-based approaches to reduce the bag size, we adopt a compression-based method to ensure that the class-relative features are preserved during compression.

Cross-Bag Augmentation enhances the input bag through three aspects: instance-level representation (Cross-Bag Augmentation), bag size increase (Instance Expansion), and bag size decrease (Instance Compression). 
Unlike recent instance sampling-based augmentation methods~\cite{psemix, rankmix, remix}, our compression-based approach preserves the class-relevant features.



\subsection{Bag-level Contrastive Learning}
\label{sec:bag-level}

We propose a bag-level contrastive learning approach to bring semantically similar bag representations closer to each other in the embedding space. 
We adopt two branches to perform bag-level contrastive learning, where the parameters of the teacher branch are updated by the student model using EMA, similar to DINO~\cite{dino}. 
Specifically, for a given input bag $X_i$, we pass $X_i$ through two branches (student, teacher) using different augmentations and generate two bag representations $z_{i}$ and $z_{i}^{'}$.
The bag representation is obtained by feeding the MIL model output into the projection head, followed by L2 normalization~\cite{dino, moco}. 
The positive sample $ z_i^+ $ is generated by the momentum encoder, which serves as the teacher model. 
The negative samples $ z_{i}^{-} $ are the bag representations from other bags in the teacher branch and are stored in a memory bank. 
The memory bank operates in a first-in, first-out manner with capacity $k$.
The loss function is formulated as:
{\small
\begin{equation}
        L_{bag} = -\log \frac{\exp(\text{sim}(z_i, z_i^+) / \tau)}{\sum_{j=1}^{k + 1} \exp(\text{sim}(z_{i}, z_{j}) / \tau)}
\end{equation}
}
where $\text{sim}(\cdot)$ denotes the cosine similarity function, $z_j$ is the sum of one positive sample and $k$ negative samples, and $\tau$ is the temperature parameter.

\subsection{Group-level Contrastive Learning}
\label{sec:group-level}
Directly performing instance-level contrastive learning may introduce training noise, as instances from different bags can share similar semantic meanings, yet they are treated as negative sample pairs.
Therefore, we propose grouping instances with similar semantics into the same group and performing group-level contrastive learning~\cite{wang2023retccl_cross_level_group, wang2021unsupervised_cross_level} to reduce training noise.
We propose Group-level Contrastive Learning to push instances with similar sematic feature closer in the embedding space.
We denote the instance outputs of the student and teacher MIL models as $ H_{n_s}\in \mathbb{R}^{n_s \times d} $ and $ H_{n_t} \in \mathbb{R}^{n_t\times d} $, respectively, where $n_s$ and $n_t$ are the number of instances for the student and teacher models.
Since $n_s$ and $n_t$ are not always equal, it is difficult to directly perform group-level contrastive learning.
We perform bag size alignment by compressing $H_{n_s}$ and $H_{n_t}$ into the same shape to enable group-level contrastive learning.

To perform size alignment, we initialize $C$ learnable prototypes and compute the cosine similarity matrices 
The similarity matrix indicates the cosine similarity between each instance and prototype pair, totaling $n \times C$ pairs.
For each instance, we determine its most similar prototype based on the similarity score and assign the instance to that prototype. 
This results in $C$ groups, where each group contains one prototype 
and the corresponding instances 
assigned to it.
This procedure applies to both branch.
After obtaining each group, we adopt Cross Attention~\cite{chen2021crossvit_crossattention} module to compress the similar instances into prototype.
We treat each prototype as the input query $Q$; the keys $K$ and values $V$ are the instances assigned to the given prototype based on the cosine similarity score.
If a group contains only the prototype, we use the prototype as the query, keys, and values for the Cross-Attention.
The outputs of Cross Attention from both branches are $g_{s} \in \mathbb{R}^{C \times d}$ and $g_{t} \in \mathbb{R}^{C \times d}$.
It should be noted that the number of most similar instances within each group is not equal before the size alignment.

After size alignment, we obtain $C$ compressed group-level representations $g_s$ and $g_t$.
Then we apply a linear projection and L2 normalization~\cite{moco, dino} to improve the discriminability of each instance.
For the $c$-th group representation in the teacher branch, $g_t^{c} \in \mathbb{R}^{d}$, we perform a centering and sharpening operation to avoid representation collapse~\cite{dino}.
For the $c$-th group representations from the teacher and student branches, $g_s^{c}$ and $g_t^{c}$, we assume these two representations should be as close as possible. 
Therefore, we first convert both group representations into probability distributions as:
{\small
\begin{equation}
    P_{s, i}^{c} = \frac{\exp(g_{s, i}^c / \tau)}{\sum_{i=1}^{d} \exp(g_{s, i}^{c}/\tau)}
\end{equation}
\begin{equation}
    P_{t, i}^{c} = \frac{\exp(g_{t, i}^c / \tau)}{\sum_{i=1}^{d} \exp(g_{t, i}^{c}/\tau)}
\end{equation}
}
%
%
\begin{table*}[!t]
\centering
\vspace{-1.5em}
\resizebox{\textwidth}{!}{
\begin{tabular}{c|c|ccc|ccc|ccc}
\hline
\multirow{2}{*}{\textbf{MIL Models}} & \multirow{2}{*}{\textbf{Augmentation}} & \multicolumn{3}{c|}{\textbf{CAMELYON-16}}                                     & \multicolumn{3}{c|}{\textbf{TCGA-LUNG}}                                       & \multicolumn{3}{c}{\textbf{TCGA-BRCA}}                                          \\ 
                                   &                                         & \textbf{ACC}                   & \textbf{AUC}                    & \textbf{F1}                     & \textbf{ACC}                    & \textbf{AUC}                    & \textbf{F1}                     & \textbf{ACC}                    & \textbf{AUC}                    & \textbf{F1}                     \\ 
\hline
\multirow{7}{*}{\textbf{DSMIL}}    & vanilla                               & $85.6_{0.9}$                   & $87.3_{1.4}$                    & $81.5_{1.8}$                    & $88.3_{1.1}$                    & $91.1_{1.5}$                    & $85.4_{1.3}$                    & $87.1_{1.2}$                    & $89.7_{1.6}$                    & $86.1_{2.0}$                    \\
                                   & w/ RankMix                            & $86.4_{1.1}$                   & $89.0_{1.3}$                    & $84.7_{1.7}$                    & $87.3_{1.2}$                    & $90.4_{1.4}$                    & $85.9_{1.4}$                    & $86.8_{1.3}$                    & $89.9_{1.5}$                    & $85.4_{1.9}$                    \\
                                   & w/ MixupMIL                           & $86.2_{1.2}$                   & $88.7_{1.5}$                    & $84.5_{1.6}$                    & $88.1_{1.3}$                    & $90.2_{1.3}$                    & $84.7_{1.5}$                    & $86.6_{1.4}$                    & $89.7_{1.4}$                    & $85.2_{1.8}$                    \\
                                   & w/ AugDiff                             & $87.0_{1.3}$                   & $89.9_{1.1}$                    & $85.6_{1.5}$                    & $87.8_{1.1}$                    & $91.4_{1.1}$                    & $86.9_{1.3}$                    & $87.4_{1.2}$                    & $91.1_{1.3}$                    & $86.4_{1.7}$                    \\
                                   & w/ DPBAug                            & $86.7_{1.2}$                   & $89.4_{1.2}$                    & $85.1_{1.4}$                    & $88.5_{1.2}$                    & $90.7_{1.2}$                    & $86.4_{1.3}$                    & $87.0_{1.3}$                    & $90.5_{1.2}$                    & $85.9_{1.8}$                    \\
                                   & w/ PRDL                               & $87.0_{1.3}$                   & $89.9_{1.1}$                    & $85.6_{1.5}$                    & $89.7_{1.1}$                    & $92.1_{1.2}$                    & $87.8_{1.3}$                    & $87.3_{1.2}$                    & $91.0_{1.3}$                    & $86.3_{1.7}$                    \\
                                   & w/ $\text{C}^{2}$Aug (Ours)                  & $\mathbf{89.1_{1.2}}$          & $\mathbf{93.2_{1.9}}$           & $\mathbf{87.0_{1.0}}$           & $\mathbf{91.2_{2.0}}$           & $\mathbf{93.5_{1.7}}$           & $\mathbf{89.8_{1.3}}$           & $\mathbf{89.4_{0.6}}$           & $\mathbf{93.7_{1.9}}$           & $\mathbf{88.4_{1.1}}$           \\
\hline
\multirow{7}{*}{\textbf{TransMIL}} & vanilla                   & $86.3_{0.8}$                   & $89.9_{1.4}$                    & $85.2_{1.6}$                    & $87.6_{1.2}$                    & $91.3_{1.3}$                    & $87.1_{1.1}$                    & $83.7_{1.0}$                    & $88.4_{1.5}$                    & $83.9_{1.7}$                    \\
                                   & w/ RankMix                            & $86.8_{1.1}$                   & $90.4_{1.1}$                    & $86.1_{1.8}$                    & $88.7_{0.9}$                    & $91.9_{1.0}$                    & $86.7_{1.4}$                    & $84.0_{1.3}$                    & $89.0_{1.2}$                    & $84.0_{1.9}$                    \\
                                   & w/ MixupMIL                              & $87.2_{1.0}$                   & $90.2_{1.5}$                    & $85.9_{1.4}$                    & $88.5_{1.3}$                    & $91.7_{1.4}$                    & $86.5_{1.2}$                    & $84.8_{1.2}$                    & $89.8_{1.4}$                    & $84.8_{1.6}$                    \\
                                   & w/ Augdiff
                                   & $88.1_{1.3}$                   & $91.4_{1.0}$                    & $86.9_{1.7}$                    & $89.1_{0.8}$                    & $92.5_{1.2}$                    & $87.5_{1.5}$                    & $85.2_{0.9}$                    & $91.2_{1.1}$                    & $85.2_{1.8}$                    \\
                                   & w/ DPBAug                            & $87.7_{1.4}$                   & $90.9_{1.2}$                    & $86.4_{1.5}$                    & $88.9_{1.1}$                    & $92.1_{0.9}$                    & $87.2_{1.3}$                    & $85.5_{1.4}$                    & $90.5_{1.3}$                    & $85.5_{1.4}$                    \\
                                   & w/ PRDL                               & $88.0_{1.1}$                   & $91.2_{1.3}$                    & $86.8_{1.6}$                    & $90.0_{1.0}$                    & $93.2_{1.5}$                    & $87.4_{1.0}$                    & $86.8_{1.1}$                    & $90.8_{0.9}$                    & $85.8_{1.7}$                    \\
                                   & w/ $\text{C}^{2}$Aug (Ours)          & $\mathbf{91.9_{1.0}}$     & $\mathbf{94.7_{1.9}}$      & $\mathbf{91.0_{0.8}}$      & $\mathbf{91.2_{0.3}}$      & $\mathbf{94.1_{1.7}}$    & $\mathbf{88.7_{0.6}}$     & $\mathbf{87.6_{0.8}}$           & $\mathbf{92.5_{1.8}}$           & $\mathbf{86.4_{0.5}}$           \\
\hline
\multirow{7}{*}{\textbf{DTFD-MIL}} & vanilla                  & $87.4_{0.7}$                   & $89.5_{1.3}$                    & $87.0_{1.7}$                    & $88.4_{1.0}$                    & $92.7_{1.1}$                    & $87.2_{1.2}$                    & $84.2_{0.9}$                    & $87.9_{1.4}$                    & $83.2_{1.5}$                    \\
                                   & w/ RankMix                            & $88.4_{1.0}$                   & $90.9_{1.0}$                    & $87.4_{1.6}$                    & $89.5_{0.8}$                    & $94.8_{0.9}$                    & $91.2_{1.0}$                    & $84.4_{1.1}$                    & $88.9_{1.3}$                    & $84.4_{1.6}$                    \\
                                   & w/ MixupMIL                              & $88.2_{0.9}$                   & $90.7_{1.4}$                    & $87.2_{1.5}$                    & $90.4_{1.1}$                    & $94.7_{1.2}$                    & $90.1_{0.8}$                    & $85.2_{1.2}$                    & $88.7_{1.1}$                    & $85.2_{1.7}$                    \\
                                   & w/ AugDiff                             & $89.1_{1.2}$                   & $91.9_{0.9}$                    & $88.1_{1.4}$                    & $91.9_{0.7}$                    & $95.1_{1.3}$                    & $91.6_{0.9}$                    & $84.8_{0.8}$                    & $89.3_{1.2}$                    & $85.9_{1.3}$                    \\
                                   & w/ DPBAug                             & $88.7_{1.3}$                   & $91.4_{1.1}$                    & $87.7_{1.2}$                    & $90.7_{0.6}$                    & $94.9_{0.8}$                    & $91.4_{1.1}$                    & $86.6_{1.0}$                    & $90.1_{0.9}$                    & $85.7_{1.5}$                    \\
                                   & w/ PRDL                               & $89.0_{1.1}$                   & $91.7_{0.8}$                    & $87.9_{1.3}$                    & $91.8_{0.9}$                    & $95.0_{1.0}$                    & $91.5_{1.3}$                    & $86.7_{0.7}$                    & $91.2_{1.0}$                    & $85.8_{1.4}$                    \\
                                   & w/ $\text{C}^{2}$Aug (Ours)                  & $\mathbf{92.5_{0.9}}$          & $\mathbf{94.3_{1.0}}$           & $\mathbf{91.4_{0.8}}$           & $\mathbf{93.5_{1.9}}$      & $\mathbf{95.8_{2.0}}$  & $\mathbf{92.7_{1.7}}$  & $\mathbf{88.4_{1.8}}$           & $\mathbf{92.9_{1.6}}$           & $\mathbf{87.6_{1.9}}$           \\
\hline
\end{tabular}
}
\vspace{-0.7em}
\caption{
Comparison between $\text{C}^{2}$Aug and other augmentation methods uses ResNet50 features.
}
\vspace{-1.7em}
\label{tab:res50_std_subscript_random}
\end{table*}

Then, we apply the Kullback-Leibler (KL) divergence loss to minimize the distance between the distributions $P_{s}^{c}$ and $P_{t}^{c}$:
{\small
\begin{equation}
    L_{group} = \sum_{c=1}^{C} P_{s}^{c} \log \left( \frac{P_{s}^{c}}{P_{t}^{c}} \right)
\end{equation} 
}
where $L_{group}$ is the sum of total $C$ KL divergence losses.

\begin{table*}[t]
\resizebox{\textwidth}{!}{
\vspace{-1.5em}
\begin{tabular}{c|c|ccc|ccc|ccc}
\hline
\multirow{2}{*}{\textbf{MIL Models}} & \multirow{2}{*}{\textbf{Augmentation}} & \multicolumn{3}{c|}{\textbf{CAMELYON-16}} & \multicolumn{3}{c|}{\textbf{TCGA-LUNG}} & \multicolumn{3}{c}{\textbf{TCGA-BRCA}} \\ 
                                         &                                         & \textbf{ACC}                           & \textbf{AUC}                            & \textbf{F1}                          & \textbf{ACC}                          & \textbf{AUC}                         & \textbf{F1}                         & \textbf{ACC}                          & \textbf{AUC}                         & \textbf{F1}                         \\ 
\hline
\multirow{7}{*}{\textbf{DSMIL}} 
& vanilla       	& $92.9_{0.3}$ & $94.9_{0.4}$ & $91.7_{0.5}$ & $91.8_{0.2}$ & $94.7_{0.3}$ & $91.4_{0.4}$ & $92.2_{0.3}$ & $94.8_{0.4}$ & $91.1_{0.5}$ \\
& w/ RankMix    	& $93.4_{0.4}$ & $95.3_{0.3}$ & $92.8_{0.5}$ & $91.1_{0.3}$ & $95.2_{0.2}$ & $91.5_{0.4}$ & $93.1_{0.4}$ & $95.2_{0.3}$ & $91.9_{0.4}$ \\
& w/ MixupMIL      	& $94.3_{0.2}$ & $96.2_{0.5}$ & $92.0_{0.4}$ & $92.3_{0.4}$ & $96.0_{0.5}$ & $92.0_{0.3}$ & $92.9_{0.3}$ & $95.4_{0.2}$ & $91.7_{0.4}$ \\
& w/ AugDiff     	& $95.8_{0.5}$ & $95.7_{0.3}$ & $92.5_{0.5}$ & $93.2_{0.1}$ & $96.6_{0.2}$ & $92.5_{0.5}$ & $93.6_{0.2}$ & $96.1_{0.5}$ & $92.4_{0.3}$ \\
& w/ DPBAug     	& $94.5_{0.5}$ & $96.4_{0.4}$ & $93.2_{0.5}$ & $92.7_{0.2}$ & $96.3_{0.4}$ & $92.9_{0.2}$ & $94.5_{0.4}$ & $95.9_{0.3}$ & $92.2_{0.3}$ \\
& w/ PRDL        	& $94.6_{0.4}$ & $96.5_{0.3}$ & $91.4_{0.5}$ & $91.3_{0.2}$ & $96.5_{0.3}$ & $90.6_{0.3}$ & $93.5_{0.2}$ & $96.0_{0.4}$ & $92.3_{0.2}$ \\
& w/ $\text{C}^{2}$Aug (Ours)           & $\mathbf{96.9_{0.2}}$ & $\mathbf{98.0_{0.3}}$ & $\mathbf{95.7_{0.4}}$ & $\mathbf{94.7_{0.1}}$ & $\mathbf{97.5_{0.2}}$ & $\mathbf{93.6_{0.3}}$ & $\mathbf{94.6_{0.2}}$ & $\mathbf{96.8_{0.3}}$ & $\mathbf{93.9_{0.2}}$ \\
\hline
\multirow{7}{*}{\textbf{TransMIL}} 
& vanilla       & $93.9_{0.3}$ & $95.1_{0.5}$ & $94.4_{0.4}$ & $92.6_{0.2}$ &	$95.1_{0.3}$ & $91.4_{0.5}$ & $91.1_{0.3}$ & $94.3_{0.4}$ & $91.4_{0.5}$ \\
& w/ RankMix    & $94.5_{0.2}$ & $96.4_{0.4}$ & $93.3_{0.5}$ & $93.6_{0.4}$ &	$95.3_{0.5}$ & $91.3_{0.4}$ & $91.6_{0.2}$ & $94.9_{0.3}$ & $91.8_{0.4}$ \\
& w/ MixupMIL      & $94.2_{0.3}$ & $96.0_{0.5}$ & $93.0_{0.4}$ & $93.4_{0.5}$ &	$96.1_{0.5}$ & $92.1_{0.3}$ & $90.3_{0.4}$ & $95.7_{0.3}$ & $91.6_{0.4}$ \\
& w/ AugDiff     & $95.0_{0.5}$ & $97.1_{0.3}$ & $93.9_{0.5}$ & $94.1_{0.1}$ &	$95.8_{0.2}$ & $93.0_{0.5}$ & $92.7_{0.2}$ & $95.4_{0.5}$ & $92.5_{0.3}$ \\
& w/ DPBAug     & $94.8_{0.5}$ & $96.7_{0.4}$ & $93.7_{0.3}$ & $93.9_{0.2}$ &	$96.6_{0.4}$ & $92.7_{0.2}$ & $92.5_{0.5}$ & $96.2_{0.2}$ & $92.3_{0.2}$ \\
& w/ PRDL        & $94.9_{0.4}$ & $96.9_{0.3}$ & $93.8_{0.4}$ & $94.0_{0.2}$ &	$97.1_{0.5}$ & $92.8_{0.3}$ & $93.6_{0.3}$ & $95.3_{0.5}$ & $92.4_{0.2}$ \\
& w/ $\text{C}^{2}$Aug (Ours)          & $\mathbf{97.3_{0.2}}$ & $\mathbf{98.9_{0.3}}$ & $\mathbf{97.6_{0.4}}$ & $\mathbf{95.1_{0.1}}$ & $\mathbf{98.9_{0.3}}$ & $\mathbf{94.2_{0.2}}$ & $\mathbf{95.6_{0.2}}$ & $\mathbf{97.8_{0.3}}$ & $\mathbf{94.4_{0.1}}$ \\
\hline
\multirow{7}{*}{\textbf{DTFD-MIL}} 
& vanilla       	& $93.8_{0.3}$ & $95.5_{0.5}$ & $93.4_{0.5}$ & $92.9_{0.2}$ & $95.8_{0.4}$ & $91.7_{0.5}$ & $91.6_{0.3}$ & $94.4_{0.5}$ & $91.7_{0.5}$ \\
& w/ RankMix    	& $94.6_{0.2}$ & $96.5_{0.4}$ & $94.3_{0.5}$ & $93.7_{0.4}$ & $96.6_{0.3}$ & $91.4_{0.5}$ & $92.7_{0.4}$ & $94.2_{0.4}$ & $91.9_{0.5}$ \\
& w/ MixupMIL      	& $94.4_{0.4}$ & $96.2_{0.5}$ & $93.2_{0.5}$ & $93.5_{0.5}$ & $96.4_{0.5}$ & $92.2_{0.4}$ & $93.1_{0.5}$ & $95.3_{0.3}$ & $91.7_{0.4}$ \\
& w/ AugDiff     	& $95.3_{0.5}$ & $97.3_{0.3}$ & $95.1_{0.5}$ & $94.2_{0.1}$ & $97.0_{0.2}$ & $93.0_{0.5}$ & $93.8_{0.2}$ & $95.6_{0.5}$ & $92.6_{0.4}$ \\
& w/ DPBAug     	& $95.1_{0.5}$ & $97.1_{0.4}$ & $93.9_{0.3}$ & $94.0_{0.2}$ & $96.8_{0.4}$ & $92.7_{0.2}$ & $92.6_{0.5}$ & $96.4_{0.4}$ & $92.4_{0.2}$ \\
& w/ PRDL       	& $95.4_{0.4}$ & $97.2_{0.4}$ & $94.8_{0.5}$ & $94.1_{0.2}$ & $97.3_{0.5}$ & $92.1_{0.4}$ & $93.7_{0.4}$ & $96.1_{0.5}$ & $92.5_{0.2}$ \\
& w/ $\text{C}^{2}$Aug (Ours)           & $\mathbf{97.9_{0.1}}$ & $\mathbf{98.3_{0.3}}$ & $\mathbf{97.7_{0.4}}$ & $\mathbf{95.3_{0.1}}$ & $\mathbf{98.7_{0.2}}$ & $\mathbf{94.4_{0.1}}$ & $\mathbf{95.8_{0.2}}$ & $\mathbf{97.9_{0.3}}$ & $\mathbf{94.6_{0.1}}$ \\
\hline
\end{tabular}
}
\vspace{-0.7em}
\caption{
Comparison between $\text{C}^{2}$Aug and other augmentation methods uses Prov-Gigapath features.
}
\vspace{-2.0em}
\label{tab:gig}
\end{table*}

\subsection{Training and Inference}
During training, the final loss function is $L$:
{\small
\begin{equation}
    L = L_{cls} + 
                 \alpha L_{bag} + 
                 \beta L_{group} 
\end{equation}
}
where $\alpha$ and $\beta$ are hyperparameters that control the weights of $L_{bag}$ and $L_{instance}$.
$\text{C}^{2}$Aug is only applied during training. 
During inference, we feed the original bag into the MIL model without any augmentation. 

\section{Experiments}

\noindent\textbf{Datasets}
We evaluated our method on three public benchmarks, namely CAMELYON16, TCGA-BRCA, and TCGA-LUNG.
\textbf{CAMELYON16} ~\cite{cam16} contains 270 training slides (159 normal, 111 tumor) and 129 testing slides. 
\textbf{TCGA-LUNG} includes a total of 1046 slides, consisting of 534 TCGA-LUAD (Lung Adenocarcinoma) and 512 TCGA-LUSC (Lung Squamous Cell Carcinoma) cases. 
\textbf{TCGA-BRCA} includes two subtypes of breast cancer, 831 Invasive Ductal
(IDC) and 210 Invasive Lobular Carcinoma (ILC), 1041 in total.

\noindent\textbf{Evaluation Metrics.} 
We use Area Under Curve (AUC), image-level Accuracy (ACC), F1 score (F1) to eliminate the effect of our method. 
We perform five-fold cross-validation for our experiments.
Please refer to the supplementary material for further details.

\subsubsection{Implementation details.}
Please refer to the supplementary material for further details.
\subsection{Comparison with State-of-the-art Methods}

We compare $\text{C}^{2}$Aug with existing augmentation approaches, 
including Mixup-based methods MixupMIL~\cite{mixupmil}, RankMix~\cite{rankmix}, and DPBAug~\cite{dong2025disentangled_DPBAug}, 
and self-augmentation-based methods AugDiff~\cite{augdiff} and PRDL~\cite{25aaai}, 
on the CAMELYON-16, TCGA-LUNG, and TCGA-BRCA datasets, 
using the two feature extraction methods mentioned earlier. 
The results are illustrated in Tab.~\ref{tab:res50_std_subscript_random} and Tab.~\ref{tab:gig}.
Specifically, in Tab.~\ref{tab:res50_std_subscript_random}, on the TCGA-LUNG dataset, $\text{C}^{2}$Aug outperforms the second-highest method by 1.0\% in terms of AUC when using the DSMIL method. 
On the CAMELYON16 dataset, $\text{C}^{2}$Aug improves the AUC by 5.6\% compared to the baseline when using TransMIL as the base MIL model. 
In Tab.~\ref{tab:gig}, we achieve $\mathbf{97.6_{0.4}}$ accuracy on the CAMELYON16 dataset, an improvement of 3.4\% over the vanilla TransMIL. 
$\text{C}^{2}$Aug achieves remarkable performance compared to other methods, demonstrating the effectiveness of $\text{C}^{2}$Aug.

\subsection{Ablation Studies}
We first compare $\text{C}^{2}$Aug with other SOTA methods, followed by comprehensive ablation studies to verify its effectiveness.  
TransMIL is adopted as the default base MIL model unless otherwise specified.

\begin{table}[b]
\centering
\vspace{-1.5em}
\resizebox{\linewidth}{!}{
\begin{tabular}{c|ccc|ccc}
\hline
\multirow{2}{*}{\textbf{Model}} 
& \multicolumn{3}{c|}{\textbf{ResNet50}} 
& \multicolumn{3}{c}{\textbf{Prov-Gigapath}} \\
 & \textbf{ACC} & \textbf{AUC} & \textbf{F1} 
 & \textbf{ACC} & \textbf{AUC} & \textbf{F1} \\
\hline
$\text{C}^{2}$Aug    & $\mathbf{91.9_{1.0}}$     & $\mathbf{94.7_{1.9}}$  & $\mathbf{91.0_{0.8}}$  &  $\mathbf{97.3_{0.2}}$ & $\mathbf{98.9_{0.3}}$ & $\mathbf{97.6_{0.4}}$ \\
w/o CB   & $86.7_{1.4}$ & $90.9_{2.0}$ & $84.8_{1.2}$ & $92.1_{0.6}$ & $96.7_{0.7}$ & $93.4_{0.8}$ \\
w/o MVF  & $87.1_{1.3}$ & $91.5_{1.9}$ & $85.8_{1.1}$ & $93.5_{0.3}$ & $97.2_{0.6}$ & $94.8_{0.5}$ \\
w/o IC   & $87.6_{1.1}$ & $91.9_{2.1}$ & $86.3_{0.9}$ & $93.8_{0.5}$ & $97.6_{0.4}$ & $94.9_{0.7}$ \\
w/o IE   & $88.7_{1.2}$ & $92.5_{1.7}$ & $87.2_{0.8}$ & $94.1_{0.4}$ & $97.8_{0.5}$ & $95.4_{0.6}$ \\
\hline
\end{tabular}
}

\vspace{-0.5em}
\caption{Effects of Cross-Bag Augmentation on CAMELYON-16 dataset using TransMIL.
``CB'': Cross-Bag, ``MVF'': Multi-View Fusion, ``IC''/``IE'': Instance Compression/Expansion.}
\label{tab:effects_of_cross_bag}
\vspace{-1.5em}
\end{table}

\noindent\textbf{Effect of Cross-Bag Augmentation.}
$\text{C}^{2}$Aug introduces the Cross-Bag Augmentation module to perform bag-level augmentation.
Multi-View Fusion is designed to conduct instance-level augmentation, while Instance Compression and Expansion aims to augment the bag size.
We conduct ablation studies using TransMIL on both ResNet50 and ProV-Gigapath features to evaluate the effectiveness of Cross-Bag Augmentation, Multi-View Fusion, and Instance Compression and Expansion on the CAMELYON-16 dataset.
As shown in Tab.~\ref{tab:effects_of_cross_bag}, the absence of Cross-Bag Augmentation results in the most significant performance decline on both feature types.
Among the three components, Multi-View Fusion leads to the largest performance drop, indicating that instance-level augmentation is most crucial for the effectiveness of $\text{C}^{2}$Aug.

\noindent\textbf{Sampling Strategy of Cross-Bag Augmentation.}
Cross-Bag Augmentation module samples instances from all bags sharing the same class label, with each bag contributing an arbitrary number of instances to form a pseudo-bag.
We evaluate the impact of varying the number of bags and the number of instances in Cross-Bag Augmentation.
We first fix the number of bags from which instances are sampled—4, 16, 64, or all.  
We then evaluate the effect of the number of instances sampled per bag by comparing average and arbitrary counts.
We perform experiments using TransMIL as the MIL model on CAMELYON-16 Dataset.
As shown in Tab.~\ref{tab:sampling_strategy}, the performance of $\text{C}^{2}$Aug improves with an increasing number of bags from which instances are sampled, indicating that the bag count is crucial for $\text{C}^{2}$Aug. 
A larger number of bags enables more diverse combinations of instances, thereby capturing richer patterns.  
Moreover, random instance sampling yields greater performance gains compared to average sampling.
For example, when sampling from all bags with the same class label, AUC improves by 0.8\% using ResNet50 features and by 1.3\% using Prov-Gigapath features.
This is because average sampling is a special case of random sampling, and greater pattern diversity leads to improved performance.


\begin{table}[b]
\centering
\vspace{-1.5em}
\resizebox{\linewidth}{!}{
\begin{tabular}{c|ccc|ccc}
\hline
\multirow{2}{*}{\textbf{Methods}} &
\multicolumn{3}{c|}{\textbf{ResNet50}} &
\multicolumn{3}{c}{\textbf{Prov-Gigapath}} \\
 &  \textbf{ACC} & \textbf{AUC} & \textbf{F1} & \textbf{ACC} & \textbf{AUC} & \textbf{F1} \\
\hline
$\text{C}^{2}$Aug & $\mathbf{92.3_{1.0}}$ & $\mathbf{95.3_{1.2}}$ & $\mathbf{92.1_{0.5}}$ & $\mathbf{100.0_{0.0}}$ & $\mathbf{100.0_{0.0}}$ & $\mathbf{100.0_{0.0}}$ \\
 MVF (No Keep) & $89.1_{0.7}$ & $92.1_{1.2}$ & $88.6_{1.2}$ & $97.9_{0.6}$ & $98.5_{0.7}$ & $96.2_{0.8}$ \\
 IE (No Keep) & $89.8_{1.3}$ & $92.6_{1.9}$ & $88.9_{1.1}$ & $98.4_{0.3}$ & $98.9_{0.6}$ & $97.5_{0.5}$ \\
\hline
\end{tabular}
}
\vspace{-0.5em}
\caption{
Impact of keeping the input bag of Cross-Bag Augmentation on the CAMELYON16 Dataset.
We only report the results of Normal class to eliminate the effect of the label noise.
}
\vspace{-1.5em}
\label{tab:keep_input_bag}
\end{table}

\begin{table}[t]
\vspace{-1.5em}
\resizebox{\linewidth}{!}{

\begin{tabular}{c|c|ccc|ccc}
\hline
\multirow{2}{*}{\textbf{Bag}} &
\multirow{2}{*}{\textbf{Inst.}} &
\multicolumn{3}{c|}{\textbf{ResNet50}} &
\multicolumn{3}{c}{\textbf{Prov-Gigapath}} \\
 & &  \textbf{ACC} & \textbf{AUC} & \textbf{F1} & \textbf{ACC} & \textbf{AUC} & \textbf{F1} \\
\hline
4  & Mean & $88.4_{1.4}$ & $91.3_{2.0}$ & $88.2_{1.2}$ & $94.5_{0.6}$ & $96.3_{0.7}$ & $94.7_{0.8}$  \\
4 & Rand & $89.5_{1.4}$ & $92.3_{2.0}$ & $88.6_{1.2}$ & $94.9_{0.6}$ & $96.5_{0.7}$ & $95.2_{0.8}$ \\
16  & Mean & $89.8_{1.3}$ & $92.6_{1.9}$ & $88.9_{1.1}$ & $95.2_{0.3}$ & $96.8_{0.6}$ & $95.5_{0.5}$ \\
16  & Rand & $90.1_{1.4}$ & $92.9_{2.0}$ & $89.2_{1.2}$ & $95.5_{0.6}$ & $97.1_{0.7}$ & $95.8_{0.8}$ \\
64 & Mean & $90.4_{1.3}$ & $93.2_{1.9}$ & $89.5_{1.1}$ & $95.8_{0.3}$ & $97.4_{0.6}$ & $96.1_{0.5}$ \\
64 & Rand & $90.7_{1.4}$ & $93.5_{2.0}$ & $89.8_{1.2}$ & $96.1_{0.6}$ & $97.7_{0.7}$ & $96.4_{0.8}$ \\
All & Mean & $87.6_{1.1}$ & $93.9_{2.1}$ & $86.3_{0.9}$ & $93.8_{0.5}$ & $97.6_{0.4}$ & $94.9_{0.7}$ \\
All & Rand & $\mathbf{91.9_{1.0}}$ & $\mathbf{94.7_{1.9}}$ & $\mathbf{91.1_{0.8}}$ & $\mathbf{97.3_{0.2}}$ & $\mathbf{98.9_{0.3}}$ & $\mathbf{97.6_{0.4}}$ \\
\hline
\end{tabular}
}
\vspace{-0.5em}
\caption{
The impact of sampling strategies in Cross-Bag Augmentation: 
``Bag" is the number of bags from which instances are sampled.
``Inst." is the number of instances sampled per bag, 
``Mean" samples the same number of instances per bag; ``Rand" samples randomly.
}
\label{tab:sampling_strategy}
\vspace{-2em}
\end{table}

\noindent\textbf{Impact of Retaining the Input Bag in Cross-Bag Augmentation.}
Instance Expansion (IE) and Multi-View Fusion (MVF) retain the input bag during augmentation.  
We aim to investigate whether retaining the input bag during augmentation is necessary.  
Since Instance Expansion inherently preserves the input bag, we conduct experiments solely on IE and MVF.  
We replace the input bag with instances randomly sampled from all bags of the same class.  
Given that random sampling may introduce label noise—particularly in tumor slides—we perform experiments on the CAMELYON-16 dataset while replacing only the input bags of normal slides with randomly sampled instances.  
TransMIL is employed as the MIL model.  
As shown in Tab.~\ref{tab:keep_input_bag}, replacing the input bag in Multi-View Fusion (MVF No Keep) results in an AUC drop of 3.2\% with ResNet50 features and 1.5\% with Prov-Gigapath features.  
Replacing the input bag within Instance Expansion (IE No Keep) also leads to reduced performance.
The performance degradation on normal slides is attributed to the loss of semantic content within the input bag.
The semantic content refers to the instance distribution within a bag.  
For example, some tumor slides contain less than 1\% tumor region, and the average tumor area across tumor slides is 20\%.  
Sampling instances across all tumor bags would result in each pseudo-bag containing approximately 20\% tumor instances, thereby reducing the diversity of pseudo-bags.
Retaining the input bag in Cross-Bag Augmentation increases the diversity of pseudo-bags, leading to improved model performance.

\noindent\textbf{Comparison of Masking Methods.}
Multi-View Fusion generates a varying number of views for each instance by employing three masking strategies: Element-wise Random Masking (Element-Wise), Row-wise Random Masking (Row-Wise), and Top-k Random Masking (Top-K). 
Ablation studies on these three methods, alongside a fixed number of views per instance (No Mask), are conducted using the CAMELYON-16 dataset with TransMIL as the MIL model. 
As illustrated in Tab.~\ref{tab:masking_methods}, Row-wise Random Masking achieves the highest performance when using two features. 
Conversely, fixed view methods yield the worst performance due the lack of diversity compared to random view methods.
The Element-wise masking method underperforms compared to the Row-wise masking method because the number of masked views for each instance follows a binomial distribution, as illustrated in Eq.~\ref{label:element_wise_random_mask}. 
As a result, the probability of observing a large number of masked views per instance decreases, making higher view counts unlikely.
The Row-wise masking method ensures that the number of views per instance follows a uniform distribution, as illustrated in Eq.~\ref{label:row_wise_random_mask}, thereby increasing the diversity of instances.
Row-wise random masking is adopted by default for all experiments.

\begin{table}[b]
\centering
\vspace{-1.5em}
\resizebox{\columnwidth}{!}{%
\begin{tabular}{c|ccccc}
\hline
\textbf{Methods} & \textbf{$<$1\%} & \textbf{1\%$\sim$10\%} & \textbf{10\%$\sim$100\%}& \textbf{Normal} & \textbf{Total} \\
\hline
w/o $L_{group}$ + $L_{bag}$   & $80.6_{1.6}$ & $89.5_{1.3}$ & $92.8_{0.8}$ & $92.7_{1.0}$ & $89.4_{0.5}$\\
w/o $L_{group}$               & $81.1_{0.4}$ & $90.3_{0.5}$ & $93.1_{1.0}$ & $92.8_{0.5}$ & $90.1_{0.7}$\\
w/o $L_{bag}$                 & $81.3_{1.0}$ & $90.0_{1.0}$ & $93.3_{1.2}$ & $93.4_{0.3}$ & $90.4_{0.5}$ \\
$\mathrm{C}^{2}$Aug           & $\mathbf{82.4_{1.4}}$ & $\mathbf{91.4_{1.3}}$ & $\mathbf{93.4_{0.4}}$ & $\mathbf{93.8_{0.4}}$ & $\mathbf{91.9_{1.0}}$ \\
\hline
\end{tabular}%
}
\vspace{-0.5em}
\caption{Effects of $L_{bag}$ and $L_{group}$ on the CAMELYON-16 Dataset.}
\label{tab:four_group_less_than_1percent}
\vspace{-1.5em}
\end{table}

\noindent\textbf{Effects of $L_{bag}$ and $L_{group}$.}
To evaluate the effectiveness of $L_{bag}$ and $L_{group}$, the CAMELYON-16 test set, comprising 129 bags, is divided into four groups based on tumor instance ratio: healthy bags (81 bags), and tumor bags with tumor instance ratios below 1\% (23 bags), between 1\% and 10\% (17 bags), and above 10\% (8 bags), respectively.
We report the accuracy on the four groups, as shown in Tab.~\ref{tab:four_group_less_than_1percent}.
Without $L_{bag}$ and $L_{group}$, the performance degradation is most pronounced in the groups with tumor ratios below 1\% (-1.8\%) and between 1\% and 10\% (-1.9\%).
This indicates that bag-level and group-level contrastive learning can enhance performance on challenging instances, particularly those with small tumor instance ratios.

\begin{table}[t]
\centering
\vspace{-1.5em}
\resizebox{\linewidth}{!}{
\begin{tabular}{c|ccc|ccc}
\hline
\multirow{2}{*}{\textbf{Inst.}} &
\multicolumn{3}{c|}{\textbf{ResNet50}} &
\multicolumn{3}{c}{\textbf{Prov-Gigapath}} \\
 &  \textbf{ACC} & \textbf{AUC} & \textbf{F1} & \textbf{ACC} & \textbf{AUC} & \textbf{F1} \\
\hline
No Mask & $89.4_{1.2}$ & $92.2_{1.7}$ & $88.6_{1.1}$ & $94.8_{0.9}$ & $96.4_{0.6}$ & $95.0_{1.3}$ \\
Top-k & $90.8_{0.7}$ & $93.6_{1.4}$ & $90.1_{1.0}$ & $96.2_{1.1}$ & $97.8_{0.5}$ & $96.5_{0.9}$ \\
Element-Wise & $90.3_{1.3}$ & $93.0_{1.2}$ & $89.4_{0.8}$ & $95.6_{1.4}$ & $97.2_{1.0}$ & $95.9_{1.2}$ \\
$\text{C}^{2}$Aug & $\mathbf{91.9_{1.0}}$ & $\mathbf{94.7_{1.9}}$ & $\mathbf{91.1_{0.8}}$ & $\mathbf{97.3_{0.2}}$ & $\mathbf{98.9_{0.3}}$ & $\mathbf{97.6_{0.4}}$ \\
\hline
\end{tabular}
}
\vspace{-0.5em}
\caption{Comparison of Different Masking Methods in Multi-View Fusion on the  CAMELYON-16
Dataset.}
\vspace{-1em}
\label{tab:masking_methods}
\end{table}

\begin{figure}[!t]
  \centering
  \vspace{-0.3em}
  \includegraphics[width=\linewidth]{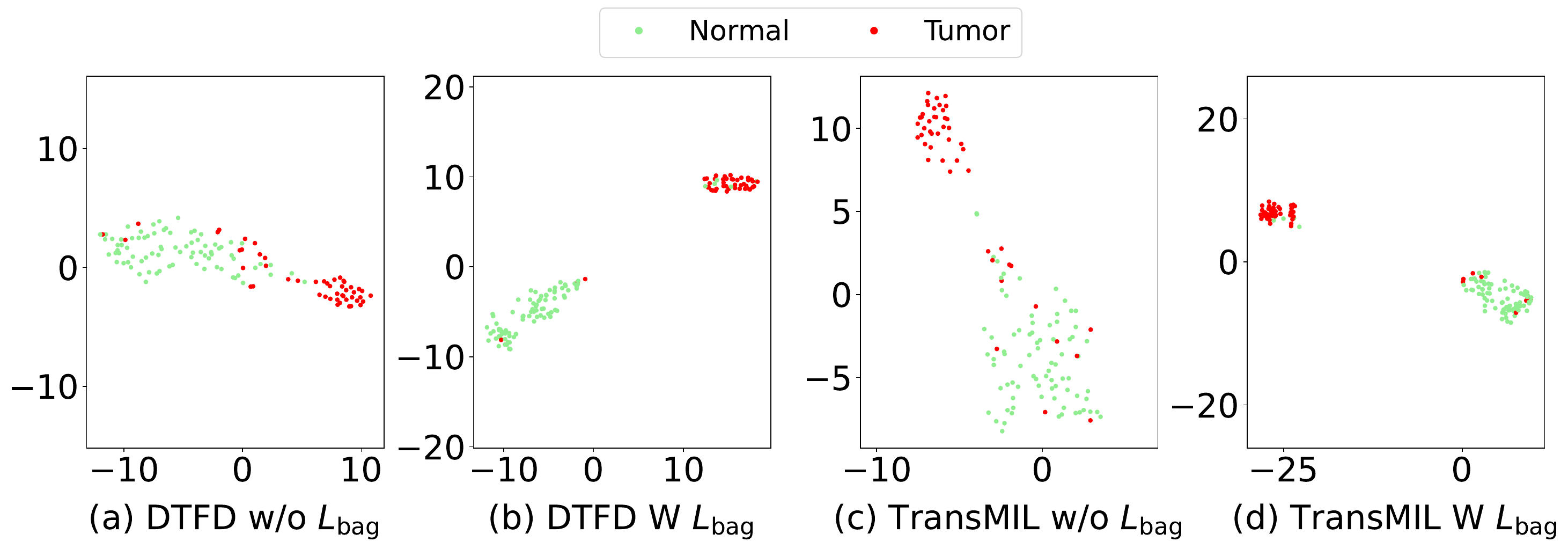}
  \vspace{-1.5em}
  \caption{
    The tSNE~\cite{maaten2008visualizing_tsne} visualization of bag-level features from the CAMELYON-16 dataset is presented. 
    It compares (a) and (b), which depict the DTFD model trained on ResNet50 extracted features without and with $L_{bag}$, respectively.
    Similarly, (c) and (d) show the Transmil model trained on ResNet50 extracted features without $L_{group}$ and with $L_{bag}$, respectively.
  }
  \label{fig:vis_bag_level}
  \vspace{-1.8em}
\end{figure}


\begin{figure}[!t]
  \centering
  \vspace{0.5em}
  \includegraphics[width=\linewidth]{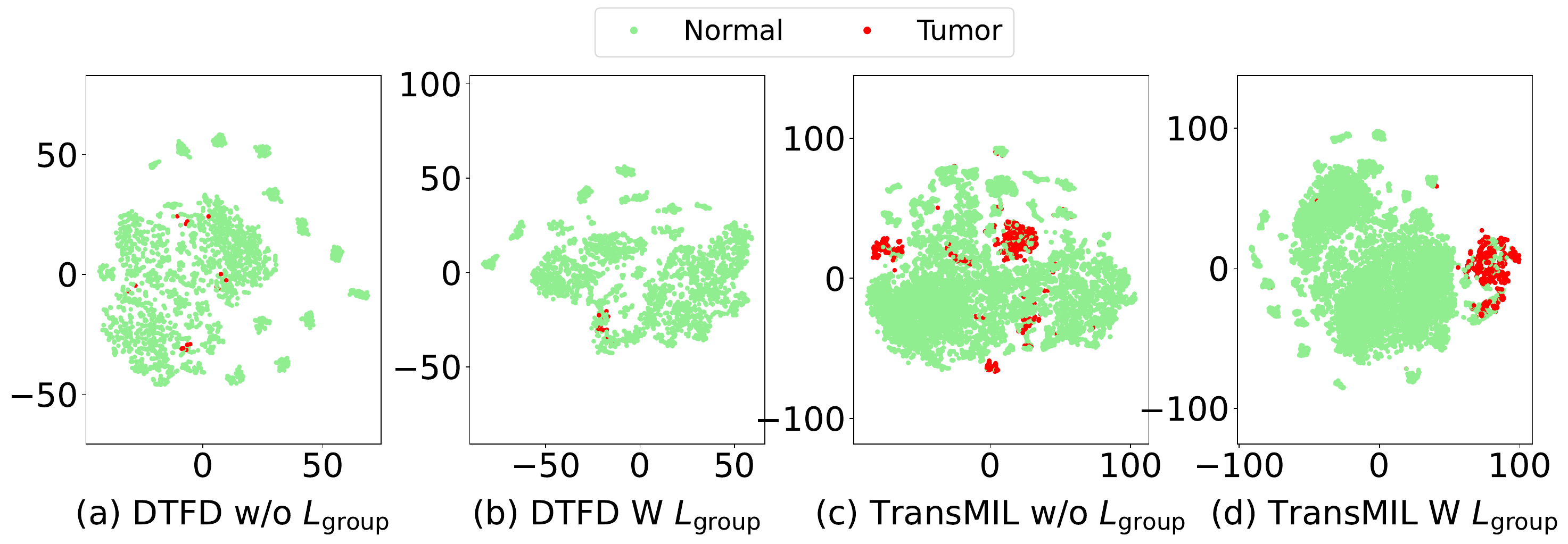}
  \vspace{-1.5em}
  \caption{
    The tSNE visualization compares instance-level features from the CAMELYON-16 dataset for different models and conditions: 
    (a) DTFD without $L_{group}$, 
    (b) DTFD with $L_{group}$, 
    (c) TransMIL without $L_{group}$, and 
    (d) TransMIL with $L_{group}$, all using features extracted by ResNet50.
  }
  \label{fig:vis_instance}
  \vspace{-1.8em}
\end{figure}

\noindent \textbf{Tsne Visualization of $L_{bag}$ and $L_{group}$.}
We visualize the bag-level representations of the test set from the CAMELYON-16 dataset, trained with and without $L_{bag}$, to demonstrate their effectiveness.
As illustrated in Fig.~\ref{fig:vis_bag_level}, the bag-level representations trained without $L_{bag}$ exhibit a more uniform distribution of tumor features.
In contrast, when trained with $L_{bag}$, the tumor bag-level features tend to cluster together.
This demonstrates that $L_{bag}$ encourages features with similar semantic content to become closer in the embedding space.
We also performed t-SNE visualization of instance-level features trained with and without $L_{group}$, as shown in Fig.~\ref{fig:vis_instance}.
Subfigures (a) and (b) correspond to the same bag, and subfigures (c) and (d) correspond to another identical bag.
After training with $L_{group}$, tumor instances are clustered together, even when the number of tumor instances is small.
Therefore, $L_{bag}$ and $L_{group}$ encourage bag-level and instance-level features with similar semantics to become closer in the embedding space, thereby enhancing the discriminative capability of the MIL models.

\section*{Conclusion}
We propose $\text{C}^2$Aug to sample instances from all bags sharing the same class label as the input bag to increase the diversity of generated pseudo-bags.  
$\text{C}^2$Aug incorporates three augmentation methods to perform instance-level and bag-size transformations.  
It further introduces Bag-level and Group-level Contrastive Learning to push features with similar semantic meanings closer in the embedding space.  
$\text{C}^2$Aug preserves all instances from the input bag, ensuring that no critical instances are discarded, in contrast to recent selection-based augmentation methods.  
Extensive experiments demonstrate the effectiveness of $\text{C}^2$Aug.

\bibliography{aaai2026}

\begin{thebibliography}{26}
\providecommand{\natexlab}[1]{#1}

\bibitem[{Bejnordi et~al.(2017)Bejnordi, Veta, Van~Diest, Van~Ginneken, Karssemeijer, Litjens, Van Der~Laak, Hermsen, Manson, Balkenhol et~al.}]{cam16}
Bejnordi, B.~E.; Veta, M.; Van~Diest, P.~J.; Van~Ginneken, B.; Karssemeijer, N.; Litjens, G.; Van Der~Laak, J.~A.; Hermsen, M.; Manson, Q.~F.; Balkenhol, M.; et~al. 2017.
\newblock Diagnostic assessment of deep learning algorithms for detection of lymph node metastases in women with breast cancer.
\newblock \emph{Jama}, 318(22): 2199--2210.

\bibitem[{Caron et~al.(2021)Caron, Touvron, Misra, J{\'e}gou, Mairal, Bojanowski, and Joulin}]{dino}
Caron, M.; Touvron, H.; Misra, I.; J{\'e}gou, H.; Mairal, J.; Bojanowski, P.; and Joulin, A. 2021.
\newblock Emerging properties in self-supervised vision transformers.
\newblock In \emph{Proceedings of the IEEE/CVF international conference on computer vision}, 9650--9660.

\bibitem[{Chen, Fan, and Panda(2021)}]{chen2021crossvit_crossattention}
Chen, C.-F.~R.; Fan, Q.; and Panda, R. 2021.
\newblock Crossvit: Cross-attention multi-scale vision transformer for image classification.
\newblock In \emph{Proceedings of the IEEE/CVF international conference on computer vision}, 357--366.

\bibitem[{Chen et~al.(2020)Chen, Kornblith, Norouzi, and Hinton}]{chen2020simple_simclr}
Chen, T.; Kornblith, S.; Norouzi, M.; and Hinton, G. 2020.
\newblock A simple framework for contrastive learning of visual representations.
\newblock In \emph{International conference on machine learning}, 1597--1607. PmLR.

\bibitem[{Chen and Lu(2023)}]{rankmix}
Chen, Y.-C.; and Lu, C.-S. 2023.
\newblock Rankmix: Data augmentation for weakly supervised learning of classifying whole slide images with diverse sizes and imbalanced categories.
\newblock In \emph{Proceedings of the IEEE/CVF Conference on Computer Vision and Pattern Recognition}, 23936--23945.

\bibitem[{Dai et~al.(2024)Dai, Wang, Wang, Zhang et~al.}]{augdiff}
Dai, L.; Wang, Y.; Wang, H.; Zhang, Y.; et~al. 2024.
\newblock AugDiff: Diffusion based feature augmentation for multiple instance learning in whole slide image.
\newblock \emph{IEEE Transactions on Artificial Intelligence}.

\bibitem[{Dong et~al.(2025)Dong, Jiang, Jiang, Li, Cai, and Zhang}]{dong2025disentangled_DPBAug}
Dong, J.; Jiang, J.; Jiang, K.; Li, J.; Cai, L.; and Zhang, Y. 2025.
\newblock Disentangled Pseudo-bag Augmentation for Whole Slide Image Multiple Instance Learning.
\newblock \emph{IEEE Transactions on Medical Imaging}.

\bibitem[{Gadermayr et~al.(2023)Gadermayr, Koller, Tschuchnig, Stangassinger, Kreutzer, Couillard-Despres, Oostingh, and Hittmair}]{mixupmil}
Gadermayr, M.; Koller, L.; Tschuchnig, M.; Stangassinger, L.~M.; Kreutzer, C.; Couillard-Despres, S.; Oostingh, G.~J.; and Hittmair, A. 2023.
\newblock Mixup-mil: Novel data augmentation for multiple instance learning and a study on thyroid cancer diagnosis.
\newblock In \emph{International conference on medical image computing and computer-assisted intervention}, 477--486. Springer.

\bibitem[{He et~al.(2020)He, Fan, Wu, Xie, and Girshick}]{moco}
He, K.; Fan, H.; Wu, Y.; Xie, S.; and Girshick, R. 2020.
\newblock Momentum contrast for unsupervised visual representation learning.
\newblock In \emph{Proceedings of the IEEE/CVF conference on computer vision and pattern recognition}, 9729--9738.

\bibitem[{Ilse, Tomczak, and Welling(2018)}]{abmil}
Ilse, M.; Tomczak, J.; and Welling, M. 2018.
\newblock Attention-based deep multiple instance learning.
\newblock In \emph{International conference on machine learning}, 2127--2136. PMLR.

\bibitem[{Li, Li, and Eliceiri(2021)}]{dsmil}
Li, B.; Li, Y.; and Eliceiri, K.~W. 2021.
\newblock Dual-stream multiple instance learning network for whole slide image classification with self-supervised contrastive learning.
\newblock In \emph{Proceedings of the IEEE/CVF conference on computer vision and pattern recognition}, 14318--14328.

\bibitem[{Li et~al.(2023)Li, Li, Savarese, and Hoi}]{li2023blip_qformer}
Li, J.; Li, D.; Savarese, S.; and Hoi, S. 2023.
\newblock Blip-2: Bootstrapping language-image pre-training with frozen image encoders and large language models.
\newblock In \emph{International conference on machine learning}, 19730--19742. PMLR.

\bibitem[{Liu et~al.(2024)Liu, Ji, Zhang, and Ye}]{psemix}
Liu, P.; Ji, L.; Zhang, X.; and Ye, F. 2024.
\newblock Pseudo-bag mixup augmentation for multiple instance learning-based whole slide image classification.
\newblock \emph{IEEE Transactions on Medical Imaging}, 43(5): 1841--1852.

\bibitem[{Maaten and Hinton(2008)}]{maaten2008visualizing_tsne}
Maaten, L. v.~d.; and Hinton, G. 2008.
\newblock Visualizing data using t-SNE.
\newblock \emph{Journal of machine learning research}, 9(Nov): 2579--2605.

\bibitem[{Shao et~al.(2021)Shao, Bian, Chen, Wang, Zhang, Ji et~al.}]{transmil}
Shao, Z.; Bian, H.; Chen, Y.; Wang, Y.; Zhang, J.; Ji, X.; et~al. 2021.
\newblock Transmil: Transformer based correlated multiple instance learning for whole slide image classification.
\newblock \emph{Advances in neural information processing systems}, 34: 2136--2147.

\bibitem[{Tang et~al.(2025)Tang, Jiang, Shi, Wang, Wu, and Zheng}]{25aaai}
Tang, K.; Jiang, Z.; Shi, J.; Wang, W.; Wu, H.; and Zheng, Y. 2025.
\newblock Promptable representation distribution learning and data augmentation for gigapixel histopathology WSI analysis.
\newblock In \emph{Proceedings of the AAAI Conference on Artificial Intelligence}, 7247--7256.

\bibitem[{Tang et~al.(2024)Tang, Jiang, Wu, Shi, Xie, Wang, Wu, and Zheng}]{tang2024self_ssrdl}
Tang, K.; Jiang, Z.; Wu, K.; Shi, J.; Xie, F.; Wang, W.; Wu, H.; and Zheng, Y. 2024.
\newblock Self-supervised representation distribution learning for reliable data augmentation in histopathology WSI classification.
\newblock \emph{IEEE Transactions on Medical Imaging}.

\bibitem[{Wang et~al.(2023)Wang, Du, Yang, Zhang, Wang, Zhang, Yang, Huang, and Han}]{wang2023retccl_cross_level_group}
Wang, X.; Du, Y.; Yang, S.; Zhang, J.; Wang, M.; Zhang, J.; Yang, W.; Huang, J.; and Han, X. 2023.
\newblock RetCCL: Clustering-guided contrastive learning for whole-slide image retrieval.
\newblock \emph{Medical image analysis}, 83: 102645.

\bibitem[{Wang, Liu, and Yu(2021)}]{wang2021unsupervised_cross_level}
Wang, X.; Liu, Z.; and Yu, S.~X. 2021.
\newblock Unsupervised feature learning by cross-level instance-group discrimination.
\newblock In \emph{Proceedings of the IEEE/CVF conference on computer vision and pattern recognition}, 12586--12595.

\bibitem[{Wang et~al.(2022)Wang, Xiang, Zhang, Yang, Yang, Wang, Zhang, Yang, Huang, and Han}]{wang2022scl_scl_wc}
Wang, X.; Xiang, J.; Zhang, J.; Yang, S.; Yang, Z.; Wang, M.-H.; Zhang, J.; Yang, W.; Huang, J.; and Han, X. 2022.
\newblock Scl-wc: Cross-slide contrastive learning for weakly-supervised whole-slide image classification.
\newblock \emph{Advances in neural information processing systems}, 35: 18009--18021.

\bibitem[{Yang et~al.(2022)Yang, Chen, Zhao, Yang, Zhang, He, and Yao}]{remix}
Yang, J.; Chen, H.; Zhao, Y.; Yang, F.; Zhang, Y.; He, L.; and Yao, J. 2022.
\newblock Remix: A general and efficient framework for multiple instance learning based whole slide image classification.
\newblock In \emph{International Conference on Medical Image Computing and Computer-Assisted Intervention}, 35--45. Springer.

\bibitem[{Zaffar et~al.(2023)Zaffar, Jaume, Rajpoot, and Mahmood}]{zaffar2023embedding_DA-GAN}
Zaffar, I.; Jaume, G.; Rajpoot, N.; and Mahmood, F. 2023.
\newblock Embedding space augmentation for weakly supervised learning in whole-slide images.
\newblock In \emph{2023 IEEE 20th International Symposium on Biomedical Imaging (ISBI)}, 1--4. IEEE.

\bibitem[{Zhang et~al.(2018)Zhang, Cisse, Dauphin, and Lopez-Paz}]{zhang2017mixup_mixup}
Zhang, H.; Cisse, M.; Dauphin, Y.~N.; and Lopez-Paz, D. 2018.
\newblock mixup: Beyond empirical risk minimization.
\newblock In \emph{International Conference on Learning Representations}.

\bibitem[{Zhang et~al.(2022)Zhang, Meng, Zhao, Qiao, Yang, Coupland, and Zheng}]{dtfd}
Zhang, H.; Meng, Y.; Zhao, Y.; Qiao, Y.; Yang, X.; Coupland, S.~E.; and Zheng, Y. 2022.
\newblock Dtfd-mil: Double-tier feature distillation multiple instance learning for histopathology whole slide image classification.
\newblock In \emph{Proceedings of the IEEE/CVF conference on computer vision and pattern recognition}, 18802--18812.

\bibitem[{Zhang et~al.(2025)Zhang, Gao, He, Li, and Mao}]{ZHANG2025103027_wsi_survey}
Zhang, Y.; Gao, Z.; He, K.; Li, C.; and Mao, R. 2025.
\newblock From patches to WSIs: A systematic review of deep Multiple Instance Learning in computational pathology.
\newblock \emph{Information Fusion}, 119: 103027.

\bibitem[{Zhu et~al.(2022)Zhu, Yu, Wu, Yu, Zhang, and Wang}]{zhu2022murcl_murcl}
Zhu, Z.; Yu, L.; Wu, W.; Yu, R.; Zhang, D.; and Wang, L. 2022.
\newblock MuRCL: Multi-instance reinforcement contrastive learning for whole slide image classification.
\newblock \emph{IEEE Transactions on Medical Imaging}, 42(5): 1337--1348.

\end{thebibliography}

\end{document}